%% file: acl2023.tex
\pdfoutput=1

\documentclass[11pt]{article}

\usepackage{ACL2023}

\usepackage{times}
\usepackage{latexsym}

\usepackage[T1]{fontenc}

\usepackage[utf8]{inputenc}

\usepackage{microtype}

\usepackage{inconsolata}
\usepackage{times}
\usepackage{latexsym}
\usepackage{amsmath,amssymb}
\usepackage{graphicx}
\usepackage{multirow}
\usepackage{booktabs,siunitx}
\usepackage{subcaption}
\usepackage{hyperref}
\usepackage{graphics}

\usepackage{colortbl}

\newtheorem{lemma}{Lemma}
\newtheorem{prop}{Proposition}

\usepackage{microtype}

\title{DC-MBR: Distributional Cooling for Minimum Bayesian Risk Decoding}

\author{Jianhao Yan\thanks{~ The early part of this work was done when Jianhao Yan was working at Pattern Recognition Center, Wechat AI, Tencent Inc, China.}$\,\,^{1,2}$, Jin Xu$^{3}$ , Fandong Meng$^{4}$ ,  Jie Zhou$^{4}$, Yue Zhang$^{2,5}$ \\
$^{1}$Zhejiang University, China \\
$^{2}$School of Engineering, Westlake University \\
$^{3}$Institute for Interdisciplinary Information Sciences, Tsinghua University \\
$^{4}$Pattern Recognition Center, WeChat AI, Tencent, China \\
$^{5}$Institute of Advanced Technology, Westlake Institute for Advanced Study \\
\href{mailto:elliottyan37@gmail.com}{elliottyan37@gmail.com} \\
}

\begin{document}
\maketitle
\begin{abstract}
Minimum Bayesian Risk Decoding (MBR) emerges as a promising decoding algorithm in Neural Machine Translation. 
However, MBR performs poorly with label smoothing, which is surprising as label smoothing provides decent improvement with beam search and improves generality in various tasks.
In this work, we show that the issue arises from the un-consistency of label smoothing on the token-level and sequence-level distributions.
We demonstrate that even though label smoothing only causes a slight change in the token-level, the sequence-level distribution is highly skewed. We coin the issue \emph{autoregressive over-smoothness}.
To address this issue, we propose a simple and effective method, Distributional Cooling MBR (DC-MBR), which manipulates the entropy of output distributions by tuning down the Softmax temperature.
We theoretically prove the equivalence between pre-tuning label smoothing factor and distributional cooling. 
Extensive experiments on NMT benchmarks validate that distributional cooling improves MBR in various settings. 
\end{abstract}

\section{Introduction}
Neural Machine Translation (NMT)~\cite{bahdanau2014neural,vaswani2017attention,yan2020multi} has witnessed significant progress in recent years. It models the conditional probability distribution of target language candidates given a source sentence by a using neural architecture model. Given a well-trained NMT model, the task of decoding is to select high-quality candidates according to the model distribution. The most commonly used decoding is Maximum-a-Posteriori decoding (MAP), which aims to find the most probable candidate (i.e., mode of the distribution). However, as revealed by recent studies~\cite{stahlberg-byrne-2019-nmt, yan2021rethinking}, MAP decoding can be degenerate, suffering from hallucination or being even empty.  

Minimum Bayesian Risk Decoding (MBR)~\cite{kumar2002minimum, eikema2020map} emerges as a promising alternative to MAP decoding, which seeks the candidate with the largest utility instead of the largest probability. Several advantages have been observed for MBR, such as being robust against domain shift~\cite{muller2021understanding} and avoiding beam search curse~\cite{eikema2021sampling}.
With the help of neural metrics \cite{freitag2021minimum}, MBR exceeds the \textit{de facto} MAP decoding algorithm -- beam search, achieving the state-of-the-art on several benchmarks. 

\begin{figure}[t!]
  \centering
  \includegraphics[width=0.9\linewidth]{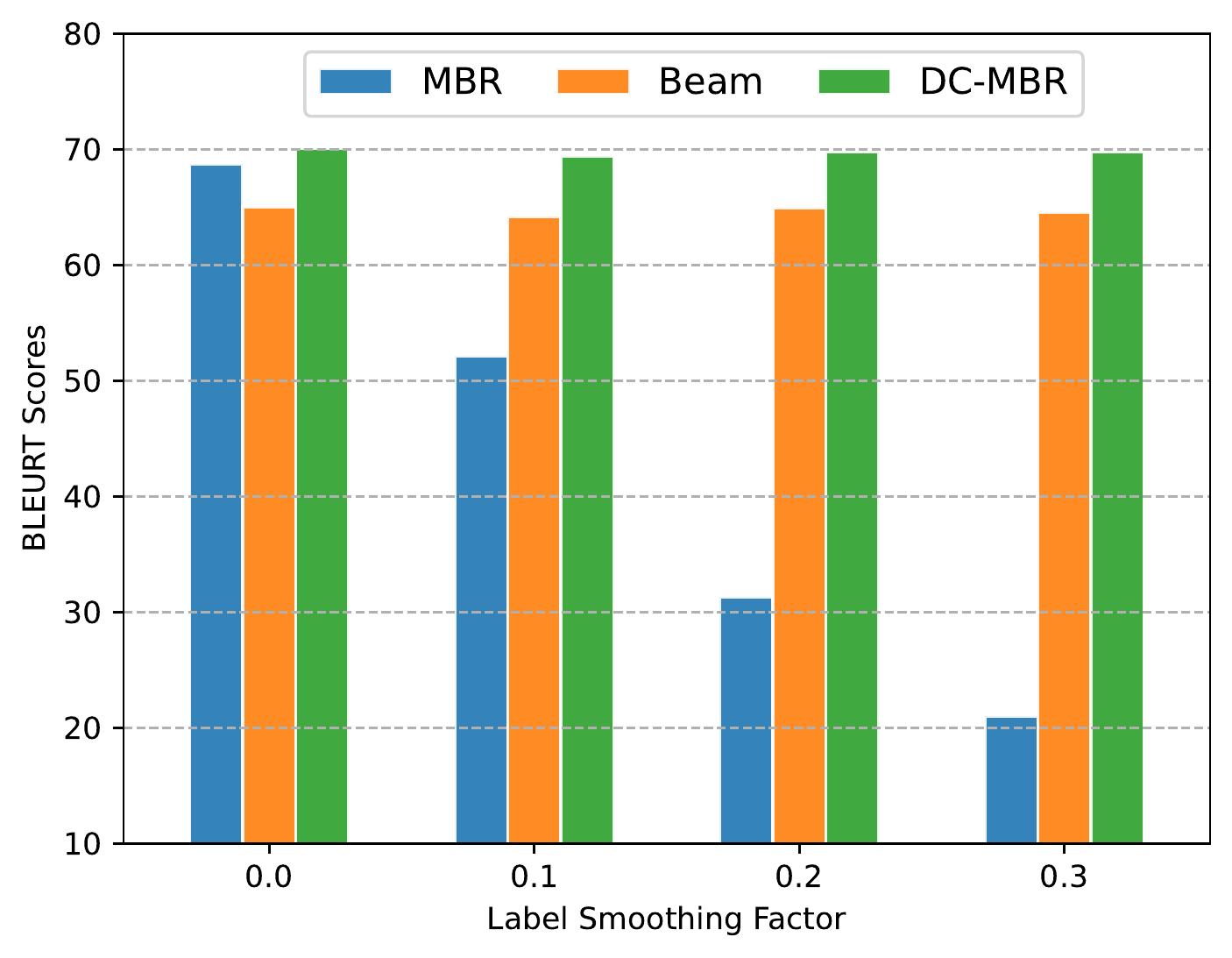}
  \vspace{-3mm}
  \caption{Translation quality against label smoothing factors. As the factor of label smoothing increases, beam search retains its performance while that of MBR drops drastically.}
\label{fig:beam_mbr_vs_ls}
\vspace{-20pt}
\end{figure}

Despite the above promises, one crucial issue is identified for MBR but not yet solved in the literature:
\emph{MBR performs poorly with models trained with label smoothing~\cite{eikema2020map}.}
We further find that the performance drops monotonically when increasing the label smoothing factor (see Figure \ref{fig:beam_mbr_vs_ls}, under the experiment settings in Section \ref{sec:exp_setup}).
It is counter-intuitive since label smoothing increases the generality of various tasks~\cite{szegedy2016rethinking, chorowski2017towards} and provides steady improvements under the MAP setting in various NMT benchmarks~\cite{vaswani2017attention,chen2018best}. 

We aim to investigate the root cause and address the issue that label smoothing benefits beam search but hurts MBR. 
As beam search makes use of token-level distribution and MBR relies on sequence-level distribution, we analyze the effect of label smoothing on the token-level and sequence-level distributions, finding that while label smoothing only slightly softens the token-level distribution, this effect makes the sequence-level distribution highly skewed with lots of low-quality candidates. We call the issue \emph{autoregressive
over-smoothness}, and quantify \emph{autoregressive over-smoothness} using token-level entropy, finding that it correlates well with MBR's performance. 

According to the above observations, we propose a conceptually simple and empirically effective approach, 
Distributional Cooling MBR (DC-MBR), which sharpens the model distributions by cooling down the Softmax temperature.
It corrects the skewed sequence-level distribution and avoids sampling from candidates that the model is not confident with. 
We theoretically prove the equivalence between distributional cooling and label (un-)smoothing, validating that distributional cooling is a reverse process of label smoothing and can safely recover from the \emph{autoregressive over-smoothness} without extra training.

We conduct experiments with two settings, bilingual NMT, under which we train Transformers \cite{vaswani2017attention} from scratch and evaluate them on three NMT benchmarks; and multilingual NMT, under which we evaluate with mBART-50~\cite{mbart-50} on ten NMT benchmarks. 
Results show that DC-MBR mitigates the autoregressive over-smoothness and significantly outperforms the de facto standard unbiased setting of MBR.
For instance, compared with naive MBR, DC-MBR improves up to 51.2 BLEURT scores for the model trained with label smoothing. 
To our knowledge, we are the first to make an in-depth investigation on MBR and label smoothing, and the first to give a principled solution. 




\section{Related Work}
\paragraph{MT Decoding}
The dominant decoding method in NMT is Maximum-a-Posteriori~(MAP) decoding, which seeks the hypothesis with the highest conditional probability.
Among all MAP decoding methods, beam search is the \textit{de facto} method in modern NMT systems. 
Many variants of beam search~\citep{bahdanau2014neural,wu2016google,he2016improved,yang2018breaking,murray2018correcting, freitag2017beam,shu2018improving} are proposed to improve its performance. 
Other than beam search, exact decoding algorithms~\cite{stahlberg-byrne-2019-nmt, yan2021rethinking} use depth-first search to find the mode or top candidates of the whole candidate space. 
However, the computational cost of exact search hinders its applications. 

Minimum Bayesian Risk Decoding (MBR), originated from SMT~\cite{kumar2002minimum, smith2006minimum, tromble2008lattice}, recently emerges as the new alternative to MAP decoding algorithm in NMT. 
MBR selects the candidates with the highest utility, e.g., an evaluation metric, instead of the highest probability, which may avoid degenerate problems with MAP. 
Early attempts to incorporate MBR into NMT mainly use the k-best list obtained via beam search~\cite{stahlberg2017neural,shu2017later,blain2017exploring}.
Recently,~\citet{eikema2020map} show that the model's sequence distribution provides a good approximation for human translation and proposes to approximate the hypothesis space and reference space of MBR by \textit{unbiased} ancestral sampling. 
This \textit{unbiased} sampling setting becomes the common practice of MBR and shows promising results.
\citet{muller2021understanding} show that MBR increases robustness against copy noise and domain shift.
\citet{eikema2021sampling} demonstrate that MBR does not suffer from beam search curse~\cite{koehn2017six}, i.e., better search always leads to better translations, and explores approximations for the expected utility.~\citet{freitag2021minimum} propose to combine neural reference-based metric (i.e., BLEURT) as the utility function and demonstrate significant improvements. 
In this work, we take the inferior performance of MBR with label smoothing as a clue and find that the \textit{unbiased} setting is sub-optimal, as MBR can be further improved by cooling down the model distribution.

\paragraph{Label smoothing}
First introduced by \citet{szegedy2016rethinking}, label smoothing is designed to improve the generality of neural models by replacing the one-hot targets with smoothed targets. 
It has shown to be effective in various NLP tasks~\cite{szegedy2016rethinking,chorowski2017towards,pereyra2017regularizing}, and provides a steady performance gain in machine translation~\cite{vaswani2017attention, chen2018best}. 
\citet{muller2019does} first study the effectiveness of label smoothing with beam search and attribute its effectiveness to better calibrating model predictions~\cite{guo2017calibration}.
However, label smoothing is not always helpful. 
\citet{meister2020generalized} observe the case where higher entropy is detrimental to the performance of random sampling. 
Here we focus on label smoothing's negative effect on MBR and attribute the issue to the different behavior of label smoothing in sequence-level and token-level distribution. Further, we introduce distributional cooling that effectively resolves this issue.



\section{Background}
We take the standard Transformer~\cite{vaswani2017attention} as the baseline, investigating label smoothing under the MAP and MBR decoding algorithms. 

\subsection{NMT and Label Smoothing}
Given a model $f(\theta)$, Neural Machine Translation (NMT) predicts the conditional probability $P(y|x)$ of a target sentence $y$ given a source sentence $x$, which can be factorised with an auto-regressive process:
\begin{equation}
    P(y|x; \theta) = \prod_{t} P(y_t|y_{<t}, x ; \theta).
\end{equation}

We refer to $P(y_t|y_{<t}, x ; \theta)$ and $P(y|x; \theta)$ as token-level distribution and sequence-level distribution, respectively. 
The token-level distribution $P(y_t|y_{<t}, x ; \theta)$ is derived with a Softmax function,
\begin{gather}
    o^i_t = f(y^i| y_{<t}, x; \theta),\\
    P(y^i_t|y_{<t}, x ; \theta) = \frac{\exp{{o^i}}}{\sum_j^{|V|} \exp{{o^j}}},
\end{gather}
where $y^i$ is $i$-th token in the vocabulary $V$, and $o$ represents the output logits.
The widely used objective for training an NMT model is the label-smoothed cross-entropy loss, defined as,
\begin{align}
    \mathbf{L}_{ls} = - \sum_{i} Q^i_{\lambda} \cdot \log {{P(y^i_t)}}.
\end{align}

$Q_{\lambda}$ is the $\lambda$-smoothed target distribution and its probability of the $i$-th token can be expressed as,
\begin{align}
    Q^i_{\lambda} = \begin{cases}
      1-\lambda & \text{if $y^i$ is golden token}\\
      \frac{\lambda}{|V|-1} & \text{otherwise}
    \end{cases}.
\end{align}


\subsection{Decoding Algorithms}
Given a model and an input, decoding algorithms select high-quality candidates from $P(y|x)$.

\noindent \textbf{\textbf{Maximum a Posteriori (MAP)}} The standard decoding algorithm in NMT is MAP decoding, which finds the candidate with the highest sequence probability (mode of the sequence distribution). 
\begin{align}
    y^{\text{MAP}} &= \text{argmax}_y P(y|x) \\
    &= \text{argmax}_{y_1, \cdots, y_T} \prod_{t}^T P(y_t|y_{<t}, x ; \theta).
\end{align}

The exact solution of MAP is computationally costly due to NMT's exponentially large search space. Hence, practitioners turn to beam search, a decoding algorithm that relied on greedy token selections.

\noindent \textbf{\textbf{Maximum Bayesian Risk (MBR)}}
Recently, it has been shown that the mode of the model's sequence distribution (i.e., MAP's optimal solution) may be degenerate or even empty~\cite{stahlberg-byrne-2019-nmt,yan2021rethinking}, which makes the mode a bad target. 
In contrast, MBR~\cite{kumar2002minimum} chooses the candidate with the highest expected utilities:
\begin{align}
    y^{\text{MBR}} = \text{argmax}_{h \in \mathcal{Y}_h} \underbrace{\mathbb{E}_{r\in \mathcal{Y}_r}[u(h, r) | x, \theta]}_{=:\mu_u(h; x,\theta)},
\end{align}
where the utility function $u$ can be a certain evaluation metric measuring the similarity between a hypothesis $h$ and a reference $r$. 
The hypothesis space $\mathcal{Y}_h$ and reference space $\mathcal{Y}_r$ are sets of all possible translations. 
Clearly, the above formulation is also intractable as both spaces are prohibitively large. 
Recently,~\citet{eikema2020map} propose a sampling-based approach that approximates both spaces with the help of the model distribution. The authors argue that the model distribution is a good approximation for human translations. 
Specifically, their approach relies on finite candidates sampled from the model's distribution, 
\begin{gather}
    \mathcal{Y}_{\text{model}} \sim \prod_{t} P(y_t|y_{<t}, x ; \theta),
\end{gather}
and uses these candidates as both the pseudo references and hypotheses:
\begin{gather}
    \hat{\mu}_u(h; x,\theta) := \frac{1}{N}\sum^{\mathcal{Y}_{\text{model}}}_{r} u(h, r), \\
    \hat{y}^{\text{MBR}} = \text{argmax}_{h \in \mathcal{Y}_{\text{model}}} \hat{\mu}_u(h; x,\theta),
\end{gather}
where $N=|\mathcal{Y}_{\text{model}}|$ is the number of candidates sampled.
In practice, the choice of the utility function can be NMT n-gram matching metrics such as BLEU~\cite{papineni2002bleu}, ChrF~\cite{popovic2015chrf} or neural metrics such as BLEURT~\cite{sellam2020bleurt}.


\section{Analyses}



In this section, we will examine the cause of label smoothing negatively impacting the performance of Minimum Bayes Risk (MBR). We will also compare the effects of label smoothing on beam search, which has been shown to work well with it in previous studies~\citep{szegedy2016rethinking,chorowski2017towards,pereyra2017regularizing,vaswani2017attention}. We will begin by outlining the details of our experimental setup.

\subsection{Setup}
\label{sec:exp_setup}

For the bilingual setting, we conduct experiments on three benchmarks: WMT 2020 English-German (En-De), WMT 2020 German-English (De-En), and WMT 2016 English-Romanian (En-Ro). 
We train Transformers from scratch using the training set and evaluate on dev/test sets. 
All models are trained for 300k steps. The batch size is 32k for En-De/De-En tokens and 16k for En-Ro.
Hyper-parameters settings except label smoothing are the same as ~\citet{vaswani2017attention}. 

For the multilingual setting, we use the pre-trained mBART-50~\cite{mbart-50}, which is designed specifically for NMT and trained to support over 50 languages. We choose 10 directions of WMT16 to evaluate our method, including En$\leftrightarrow$De, En$\leftrightarrow$Cs, En$\leftrightarrow$Fi, En$\leftrightarrow$Ro, En$\leftrightarrow$Ru. The statistics of all benchmarks are listed in Appendix.

For MBR settings, our results rely on a candidate list of two sizes: low-cost ($N$=10) and high-cost ($N$=50). In the analysis part, we use the high-cost setting. 
Following \citet{freitag2021minimum}, we use BLEURT v0.2~\cite{sellam2020bleurt} as our utility function to achieve state-of-the-art performance. 
We provide results using other utility functions in Appendix \ref{sec:util_and_eval}.
For evaluation, we mainly report BLEURT scores for both the bilingual and multilingual settings, except for bilingual experiments on WMT16 En-Ro, where we report sacreBLEU as its training corpus is case insensitive and the BLEURT metric is trained on case-sensitive data. 
For more details, please refer to Appendix \ref{detail_exp_setup}.

\subsection{Un-consistency between Token- and Sequence-level Distributions}
\begin{figure}[t!]
  \centering
  \includegraphics[width=0.8\linewidth]{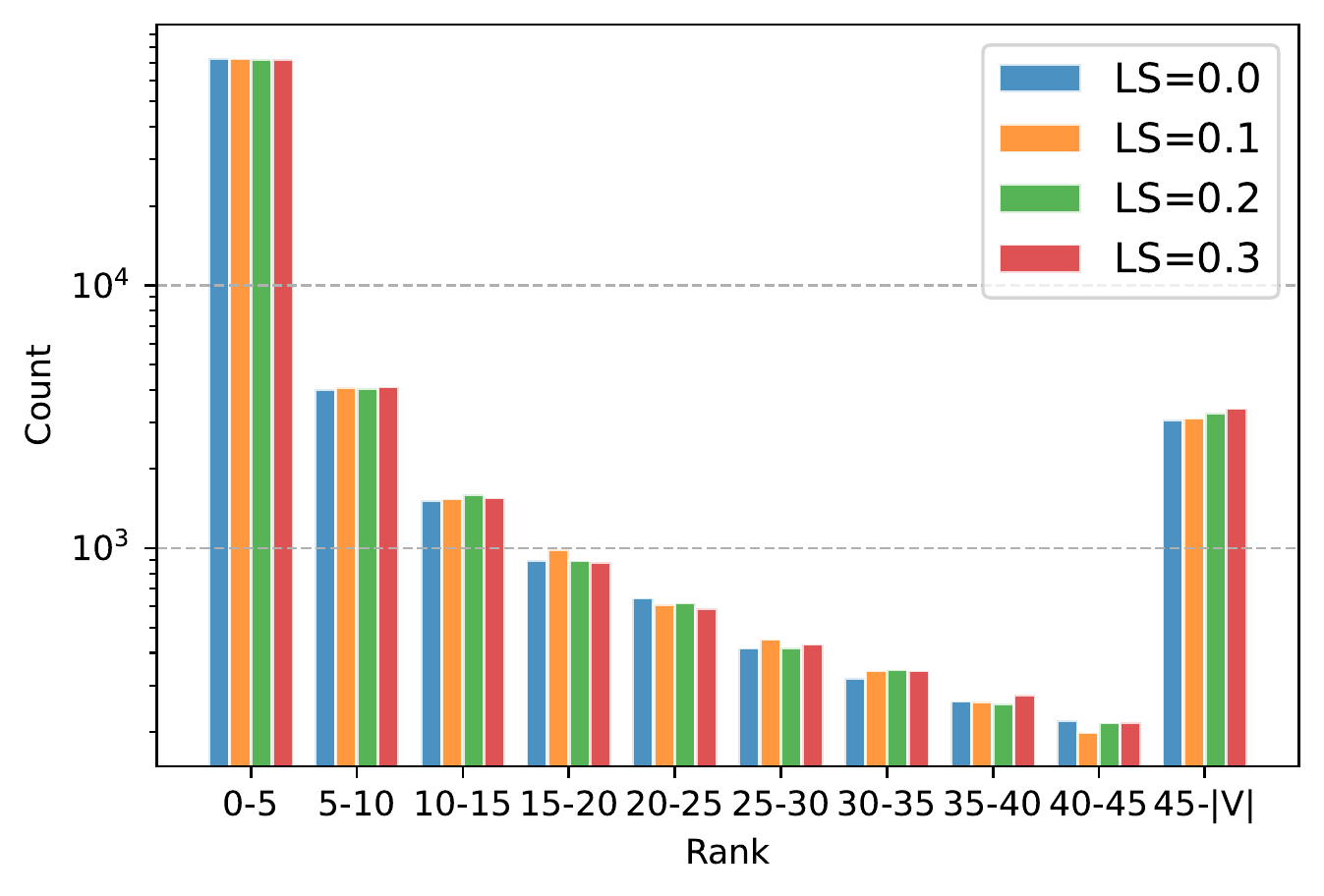} 
\caption{Ranking statistics for tokens in the ground-truth sentence within the token-level distribution $P(y_t|y_{<t}, x)$.}
\label{fig:tok_vs_ls}
\vspace{-10pt}
\end{figure}

\begin{figure}[t!]
  \centering
  \includegraphics[width=0.8\linewidth]{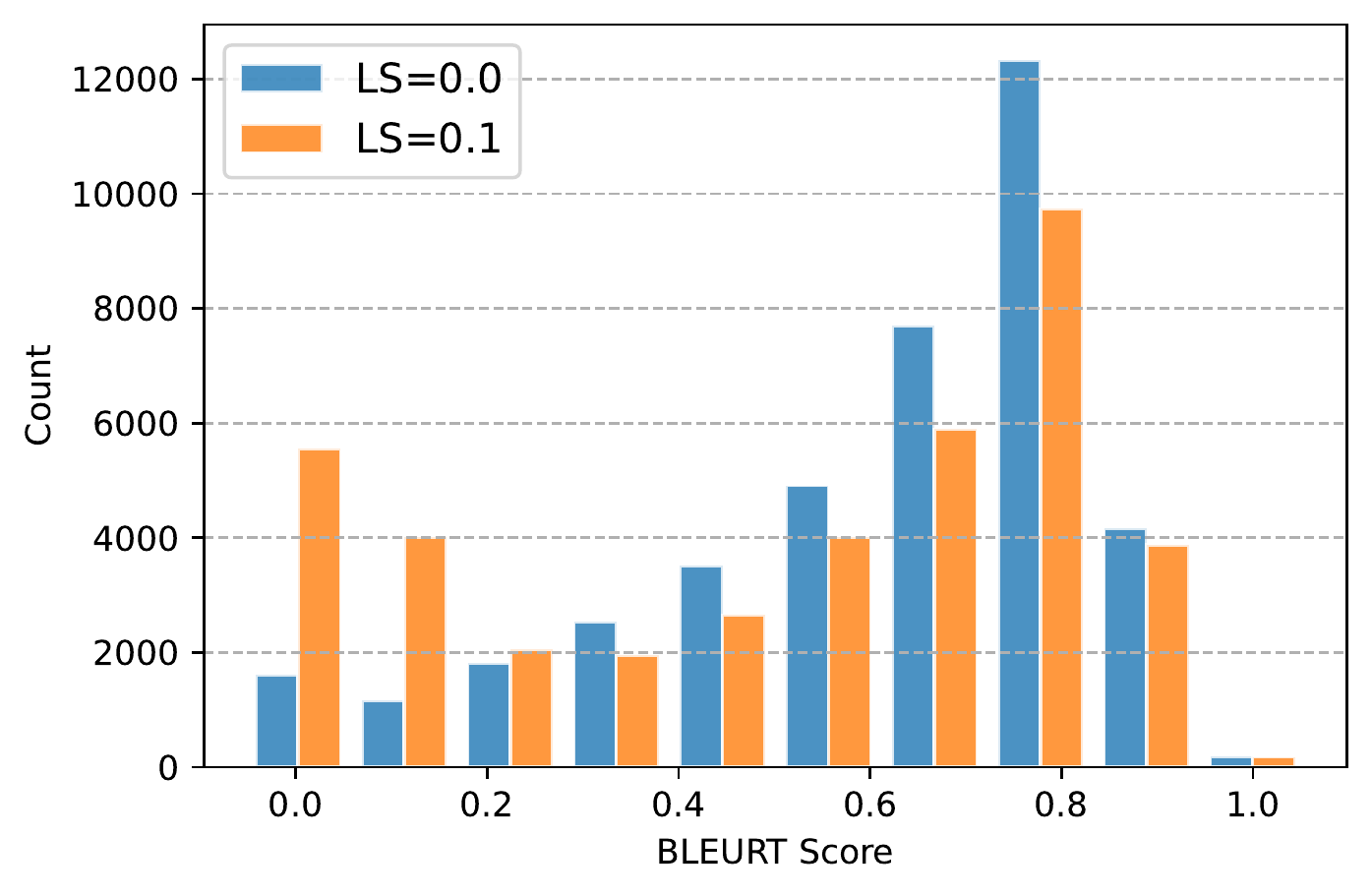}
\caption{Translation quality statistics for sequences within the top-20 candidates of the sequence-level distribution $P(y|x)$.} 
\label{fig:exact_vs_ls}
\vspace{-15pt}
\end{figure}

One key distinction between MBR and beam search is that 
MBR relies on sequence-level distribution $P(y|x)$ and re-ranks among sequence samples, while beam search generates tokens greedily from the token-level distribution $P(y_t|y_{<t}, x)$.
Therefore, we explore the effect of label smoothing over the model's sequence-level and token-level distributions. 

Figure \ref{fig:tok_vs_ls} demonstrates rankings of ground-truth tokens on the token-level distributions. 
We predict the teacher-forcing probabilities of reference tokens and investigate how the ranks of these tokens within the distribution $P(y_t|y_{<t}, x)$ change with label smoothing factors.
We can see that most of the reference tokens are ranked within the top 0-5, indicating that our models are well-trained.
When we increase the label smoothing factor, the rankings only change slightly.
Specifically, the count of reference tokens in rank 0-5 slightly drops, and that in rank 45-$|V|$ (tail of the token-level distribution) slightly increases.
{
It implies that label smoothing makes the model mildly less confident at the token-level as intended.
It improves the models' generality and accords with our experiments in Figure \ref{fig:beam_mbr_vs_ls} that label smoothing provides minor improvements with beam search.
}

The minor impact on the token-level distribution can lead to a huge disparity regarding the sequence-level distribution. 
With the exact top-$N$~\cite{yan2021rethinking} decoding algorithm, which is a DFS-based search algorithm equipped with a min-heap, we decode the topmost (i.e., top-20) sequences of the sequence-level model distribution.
Figure \ref{fig:exact_vs_ls} plots the translation qualities of these topmost sequences for models trained with and without label smoothing. 
As we can see, compared to the model without label smoothing, the model trained with label smoothing has more low-quality sequences in its top region of sequence-level distribution. 
It suggests that label smoothing skews the model distribution and gives poor sequences higher ranks/probabilities.
This may relate to the well-known label bias problem~\citep{lafferty2001conditional}, wherein the sampling process the model's short-sighted decisions on certain steps lead to poor translations. 
This leads to low-quality hypotheses and reference spaces and explains MBR's deteriorate performance in Figure \ref{fig:beam_mbr_vs_ls}.

To further understand why a small distortion in the model's token-level distribution results in a much skewed sequence-level distribution, we examine the auto-regressive nature of machine translation models. 
As a concrete example, suppose we have a reference sequence of 30 tokens. 
Given a model trained without label smoothing and \emph{perfectly} fit the data set, it should receive 100\% probability for each reference token and the whole sequence.
In contrast, with a model with label smoothing $\lambda=0.1$, each reference token receives a 90\% probability, whereas the reference sequence as a whole receives only $90\%^{30}=4\%$. 
This effect further enlarges. When $\lambda=0.2$, the reference sequence only receives about 0.1\% probability.
Consequently, as shown in Figure~\ref{fig:beam_mbr_vs_ls}, the model re-distributes the probability mass to many low-quality sentence candidates. 
We adopt the term \emph{autoregressive over-smoothness} for this issue. 
In the next section, we discuss how to quantify \emph{autoregressive over-smoothness}.

\section{Method}
We propose DC-MBR to address the above issue. The idea is to sharpen the sequence-level distribution, thus allowing the model to benefit from its own confidence.
To this end, we first propose a measure of \emph{autoregressive over-smoothness}, and then introduce our DC-MBR. 

\subsection{Measuring Autoregressive Over-Smoothness Using Entropy}
Measuring to what extent the model suffers from \emph{autoregressive over-smoothness} is non-trivial, as the search space of sequence-level distribution is prohibitively large. We turn to a token-level measure that performs well empirically, the token distribution entropy,
\begin{gather}
    H=\sum_{i}^{|V|}P (y^i_t) \log P (y^i_t).   
\end{gather}

The connection is straightforward. The lower the token level entropy is, the less probability mass is distributed to the golden reference.

\begin{figure}[t!]
  \centering
  \includegraphics[width=0.8\linewidth]{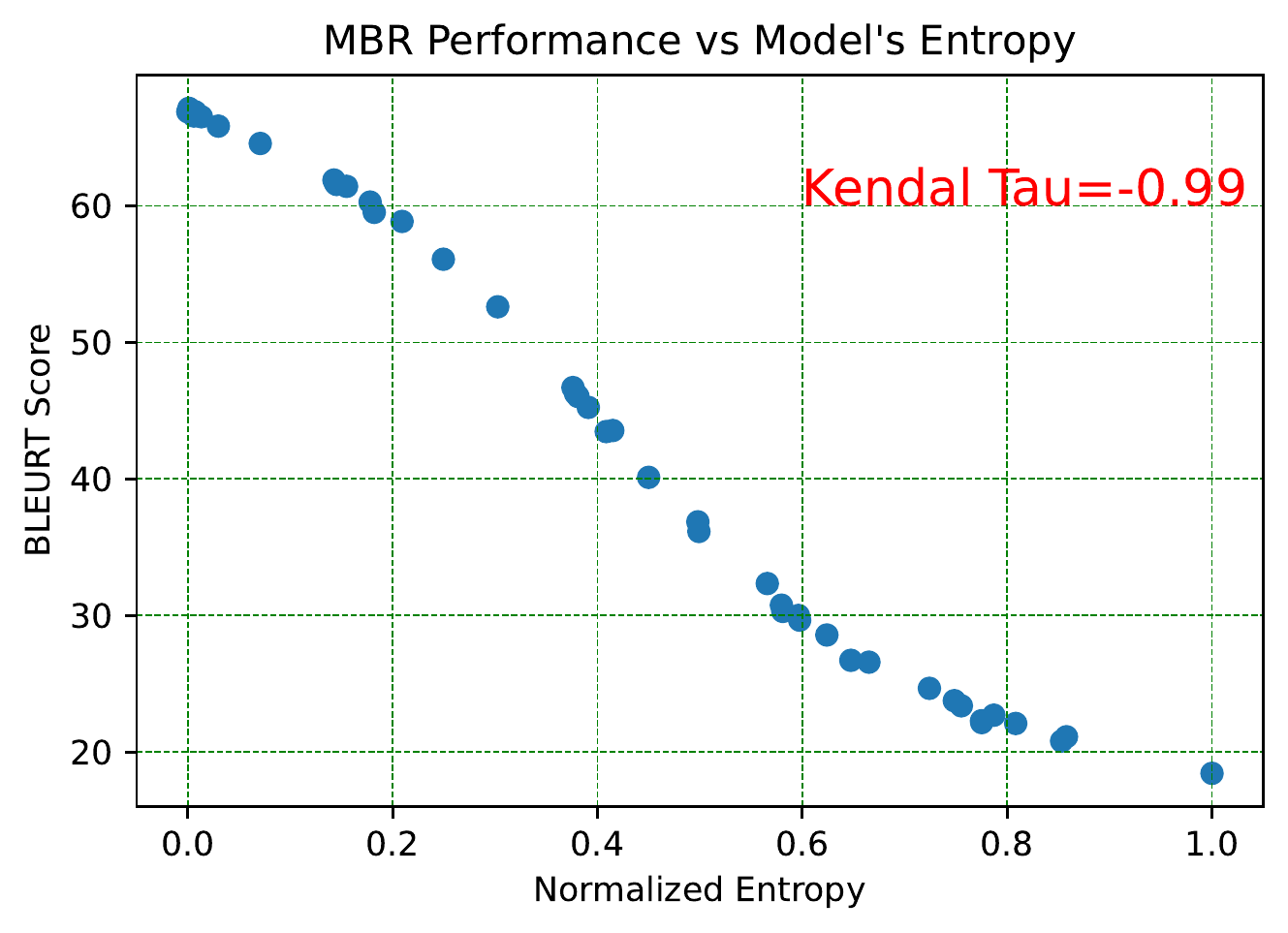}
\caption{
The performance of MBR against the smoothness of the model distribution. Each data point is a Transformer-based model. The performance of MBR is negatively correlated with entropy values. 
($\tau=-0.99$).
}
\label{fig:mbr_vs_entropy}
\vspace{-15pt}
\end{figure}

To validate our measure, we conduct experiments on the WMT20 En-De task and investigate the relationship between entropy and MBR performance. 
To control the entropy values, we train 45 Transformer-base models with various hyper-parameters (i.e., $\alpha \in [0.1, 0.9]$ and $\beta \in [0.1, 0.5]$) of generalized entropy regularizations (GER,~\citealt{meister2020generalized}), where label smoothing is one of the special cases. 
We use a Transformer-base model~\cite {vaswani2017attention} and fine-tune the model for 10k more steps.
Figure \ref{fig:mbr_vs_entropy} plots the BLEURT scores against corresponding model entropy values. 
Our measure of \emph{autoregressive over-smoothness} is a good proxy as it correlates strongly (-0.99) with the MBR performance. 
The model with smaller token distribution entropy suffers less from \emph{autoregressive over-smoothness} and achieves a better performance with MBR. 

\subsection{Distributional Cooling for MBR}
\label{sec:trmbr}


In order to mitigate the \emph{autoregressive over-smoothness} and further improve MBR, it is essential to sharpen the model distribution. This is simply the reverse of label smoothing.
While it can be achieved by training with negative entropy regularizations, it would require extra computational overhead and it would sacrifice the model's performance with beam search.

Instead, 
we manipulate the generation process of both hypothesis space and reference space by distributional cooling. It turns down the Softmax temperature and thus reduces the token-level entropy. 
Formally, we divide the logits $o$ by temperature $T$ before normalization,  
\begin{gather}
    P(y_t|y_{<t}, x ; \theta, T) = \frac{\exp{\frac{o_t}{T}}}{\sum_j^{|V|} \exp{\frac{o_j}{T}}},
\end{gather}
and generate candidates in hypothesis space and reference space with
\begin{gather}
    \mathcal{Y}^T_{\text{model}} \sim \prod_{t} P(y_t|y_{<t}, x ; \theta, T).
\end{gather}

The computation of MBR becomes
\begin{gather}
    \hat{\mu}_u(h; x,\theta) := \frac{1}{N}\sum^{\mathcal{Y}^{T_r}_{\text{model}}}_{r} u(h, r), \\
    \hat{y}^{\text{DC-MBR}} = \text{argmax}_{h \in \mathcal{Y}^{T_h}_{\text{model}}} \hat{\mu}_u(h; x,\theta).
\end{gather}

We use separate temperatures $T_h$ and $T_r$ for the hypothesis and reference spaces due to their distinct roles in MBR. See more discussion in Section \ref{sec:exp_hyp_and_ref_space}
 
With distributional cooling, we model a label (un-)smoothing process, as it forces the model to focus on its most confident candidates and avoid distributing probability mass on unconfident ones. 
It is simple and easy to implement. It only modifies the decoding phase, which can be easily applied to off-the-shelf MT models and does not affect the model's performance with beam search. 

\subsection{Proof of Equivalence}


Distributional cooling also theoretically connects with the label (un-)smoothing. In this section, we provide the proof. 


\begin{prop}
\label{prop:optimal_solution}
The optimal solution of the model trained with label smoothing $\lambda$ is $\hat{P}_{\lambda}$ whose probability of $i$-th token is:
\begin{align}
    \hat{P}^i_{\lambda} = \begin{cases}
      1-{\lambda} & \text{$y^i$ is golden token} \\
      \frac{{\lambda}}{|V|-1} & \text{otherwise}
    \end{cases}.
\end{align}
\end{prop}

Intuitively, this solution is straightforward since minimizing cross-entropy loss equals minimizing the Kullback–Leibler divergence between target distribution $Q$ and model distribution $P$. The loss achieves zero if and only if two distributions are the same. 

With the assistance of Proposition \ref{prop:optimal_solution}, we can further derive the following lemma. 
\begin{lemma}
    \label{lemma:equivalence}
    Given two models that achieve the optimal solutions with different label smoothing factors $\lambda_1, \lambda_2 < 1$, there exists a Softmax temperature factor $T=(\log \frac{1-\lambda_1}{\lambda_1})/(\log \frac{1-\lambda_2}{\lambda_2})$ that can transform $\hat{P}_{\lambda_1}$ to $\hat{P}_{\lambda_2}$.
\end{lemma}

The detailed proof for Proposition \ref{prop:optimal_solution} and Lemma \ref{lemma:equivalence} can be found in Appendix. 

The above Lemma proves the equivalence between distributional cooling and label smoothing training. 
Thus, we can exactly manipulate the Softmax temperature to recover the over-smoothness brought by label smoothing, and, furthermore, improve the performance of MBR. 
This justifies our approach in that distributional cooling with a temperature $T<1.0$ does not just make the model's output distribution sharp in any direction. It transforms the distribution towards the optimal solution of a model trained by a smaller label smoothing.



\subsection{Main Results}
\label{sec:main_results}
Table \ref{tab:en-de} shows our results on the bilingual NMT setting. Since our results on three benchmarks are consistent, we put WMT20 De-En and WMT16 En-Ro in Appendix \ref{sec:more_bi_exps}, and mainly discuss WMT20 En-De experiments in the main text. 
Table \ref{tab:multi-nmt} provides our results under the multilingual NMT setting. 
The default value of temperature is set to 0.5.     

\begin{table}[t]
\centering
\resizebox{\columnwidth}{!}{%
\begin{tabular}{c|l|c|c|c|c}
\toprule
\textbf{Models} & \multicolumn{1}{c|}{\textbf{Models}} & \textbf{BS} & \textbf{MBR} & \textbf{Ours}& $\Delta$ \\ \midrule
\multirow{4}{*}{\textbf{$N=10$}} & Transformer & 65.0 &  63.8 & 68.8 & +5.0 \\ \cline{2-6} 
                               & \cellcolor{gray!20}~~ + LS 0.1    & 64.2 & 41.0 & 67.9 & +26.9 \\ \cline{2-6} 
                               & \cellcolor{gray!20}~~ + LS 0.2    & 64.9 & 24.0 & 68.3 & +44.3 \\ \cline{2-6} 
                               & \cellcolor{gray!20}~~ + LS 0.3    & 64.6 & 17.1 & 68.3 & +51.2 \\ \midrule
\multirow{4}{*}{\textbf{$N=50$}} & Transformer & 65.0 &  68.7 & 70.1 &+1.4 \\ \cline{2-6} 
                               & \cellcolor{gray!20}~~ + LS 0.1    & 64.2 & 52.1 & 69.4& +17.3 \\ \cline{2-6} 
                               & \cellcolor{gray!20}~~ + LS 0.2    & 64.9 & 31.3 & 69.8 & +38.5 \\ \cline{2-6} 
                               & \cellcolor{gray!20}~~ + LS 0.3    & 64.6 & 21.0 & 69.8 &+48.8 \\ \bottomrule
\end{tabular}
}
\caption{BLEURT scores for En-De. Gray: Models perform poorly with original MBR. We investigate two settings: Low cost, $N$=10, 100 BLEURT calls per sentence; High cost, $N$=50, 2500 BLEURT calls per sentence. Our results are significantly better than ``MBR''($p<0.01$).}
\label{tab:en-de}
\vspace{-10pt}
\end{table}

\noindent \textbf{\emph{Mitgating Autoregressive Over-smoothness.}} 
As shown in rows 2-4 and 6-8 (i.e., gray rows) of Table \ref{tab:en-de}, we confirm that models trained by label smoothing (`\emph{+LS xx}') perform poorly with naive MBR. The performance drops drastically no matter the choice of the number of candidates or tasks. 
In contrast, DC-MBR (column \emph{Ours}) achieves strong and consistent performance across different choices of label smoothing, where the performance gap between ours and naive MBR can even reach about 50 BLEURT scores. 
This consistency indicates that our methods do not suffer from \emph{autoregressive over-smoothness}.

\noindent \textbf{\emph{Improving Sub-optimal Settings.}} 
We compare our performance with naive MBR under the unbiased setting.
In bilingual NMT (Table \ref{tab:en-de}), we observe significant gaps (+5.0/+1.4 BLEURT score) in both the low-cost scenario and the high-cost scenario. 
In multilingual NMT (Table \ref{tab:multi-nmt}), the conclusions are similar. Our method significantly outperforms MBR with +2.5 BLEURT scores when $N=10$ and with +0.5 BLEURT scores when $N=50$\footnote{Note that the significance test is based on NIST and BLEU, while the reported results here are BLEURT scores. See Appendix \ref{detail_exp_setup} for more details.}.
Our results suggest that the widely used \textit{unbiased} setting is sub-optimal, and the construction of hypothesis and reference space needs exploration.

\noindent \textbf{\emph{DC-MBR vs Beam Search.}}
Further, we compare beam search (`BS') with our approach, as beam search is the widely applied decoding algorithm in NMT applications. 
As shown in Table \ref{tab:en-de} and \ref{tab:multi-nmt}, our methods strongly outperform beam search with +4.8 to +5.1 BLEURT scores in the bilingual setting and with +1.2 and +2.2 BLEURT scores in the multilingual setting. 
Compared with naive MBR, which performs weaker than beam search when $N=10$, our methods perform much better in the low-cost scenario. 
The results indicate that our approach makes MBR more applicable in place of beam search in NMT applications, with lower costs and higher translation quality.

\noindent \textbf{\emph{Computational Cost Reduction.}} 
A by-product of our approach is our method can achieve the same performance with much less computational cost, i.e., the number of candidates, and thus enable a much faster decoding process.
For instance, our low-cost result (68.8, `Transformer, Ours, $N$=10' in Table \ref{tab:en-de}) is comparable to that of the original MBR's high-cost result (68.7, `Transformer, MBR, $N$=50'). The computational cost is reduced from 2500 to 100 BLEURT calls (25x speedup), due to the quadratic nature of MBR. 
Compared with other acceleration methods in MBR~\cite{eikema2021sampling,freitag2021minimum}, which mainly focus on truncating the hypothesis space or the reference space and accelerating the MBR computation process solely, our methods additionally reduce the cost of candidate generation. 

\begin{table}
\centering
\resizebox{\columnwidth}{!}{%
\begin{tabular}{c||c||c|c|c||c|c|c}
\toprule
 \multirow{2}{*}{\textbf{mBART}}
     & \multirow{2}{*}{\textbf{BS}}  &  \multicolumn{3}{c||}{\textbf{N=10}}      & \multicolumn{3}{c}{\textbf{N=50}} \\ \cline{3-8}
&   & \textbf{MBR}  & \textbf{Ours} & \textbf{$\Delta$} & \textbf{MBR}  & \textbf{Ours} & \textbf{$\Delta$} \\ \midrule
En-De & 71.8 & 70.3 & \textbf{73.7} & +3.4   & 74.2 & \textbf{74.8} & +0.6   \\ \midrule
De-En & 74.0 & 73.4 & \textbf{74.3} & +0.9   & \textbf{75.1} & 74.7 & -0.4  \\ \midrule
En-Cs & 73.0 & 70.1 & \textbf{75.3} & +5.2   & 75.0 & \textbf{77.2} & +2.2   \\ \midrule
Cs-En & 70.8 & 69.8 & \textbf{71.2} & +1.4   & 71.7 & \textbf{71.7} & +0.0   \\ \midrule
En-Ro & 77.2 & 77.3 & \textbf{78.6} & +1.3   & \textbf{79.8} & 79.5 & -0.3  \\ \midrule
Ro-En & 72.3 & 72.2 & \textbf{72.5} & +0.4   & \textbf{73.6} & 72.8 & -0.8  \\ \midrule
En-Ru & 70.8 & 68.3 & \textbf{72.9} & +4.6   & 72.5 & \textbf{74.4} & +1.9   \\ \midrule
Ru-En & 71.6 & 70.3 & \textbf{72.2} & +1.9   & 72.4 & \textbf{72.7} & +0.3   \\ \midrule
En-Fi & 77.4 & 75.6 & \textbf{79.9} & +4.2   & 80.1 & \textbf{81.8} & +1.7   \\ \midrule
Fi-En & 68.9 & 68.1 & \textbf{69.6} & +1.5   & 70.1 & \textbf{70.2} & +0.1   \\ \midrule
Average   & 72.8 & 71.5 & \textbf{74.0} & +2.5   & 74.5 & \textbf{75.0} & +0.5  \\ \bottomrule
\end{tabular}}
\caption{BLEURT scores for the ten tasks on WMT16 with mBART. Our results are significantly better than ``MBR'' ($p<0.01$).}
\label{tab:multi-nmt}
\end{table}


\subsection{The Number of Candidates}
In our approach, we decrease the Softmax temperature to sharpen the token-level distribution. This may reduce the diversity of generated candidates.
Thus, one possible concern is whether our method would limit the potential of MBR when using a large number of candidates. 
To this end, we study MBR's performance as the number of candidates increases. We plot different temperature choices and report the corresponding BLEURT scores over WMT20 En-De, and present results with both models trained with and without label smoothing.

\begin{figure}
\centering
\begin{subfigure}{0.48\linewidth}
  \centering
  \includegraphics[width=\linewidth]{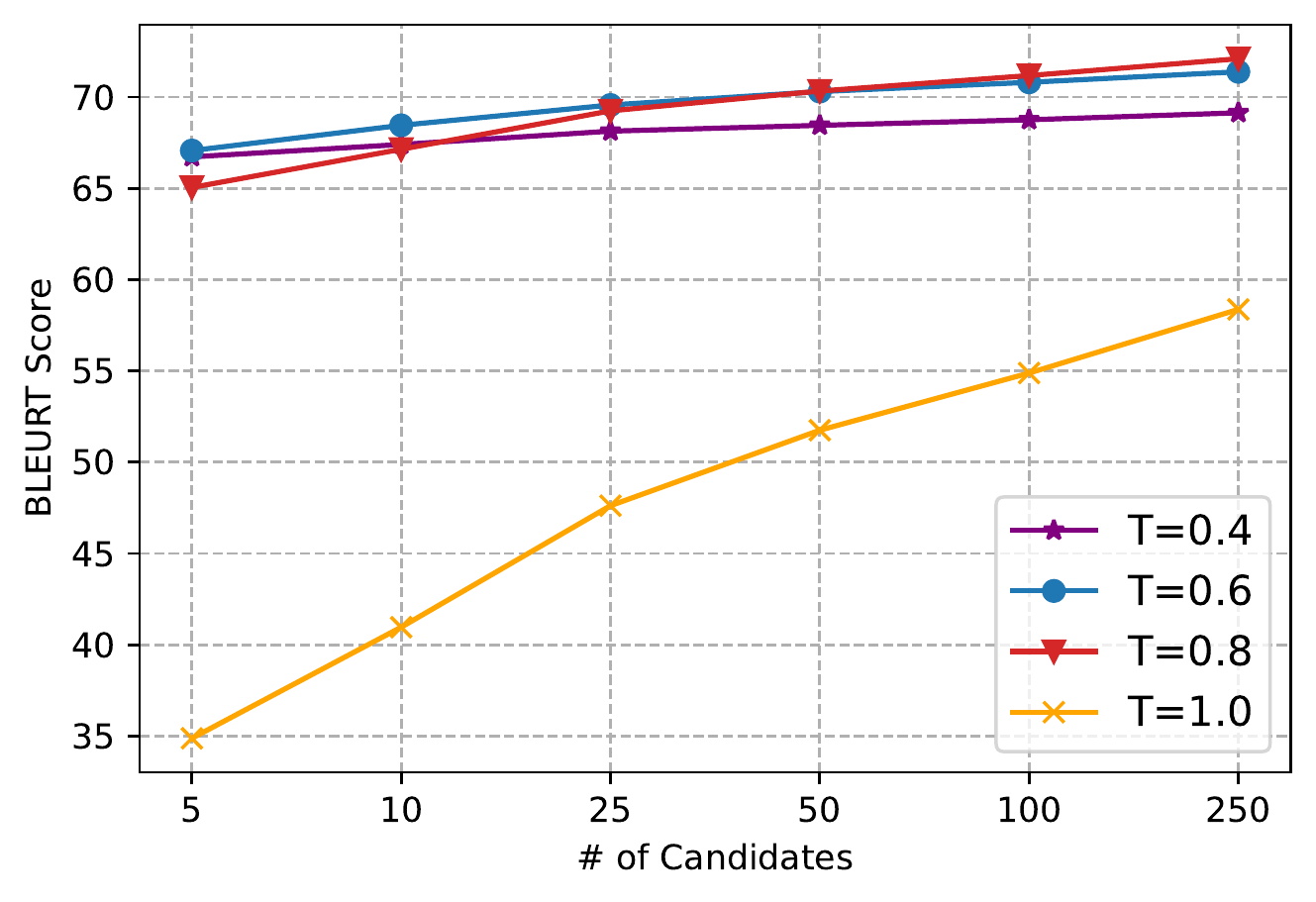}
  \label{fig:sub1}
\end{subfigure}%
\begin{subfigure}{0.48\linewidth}
  \centering
  \includegraphics[width=\linewidth]{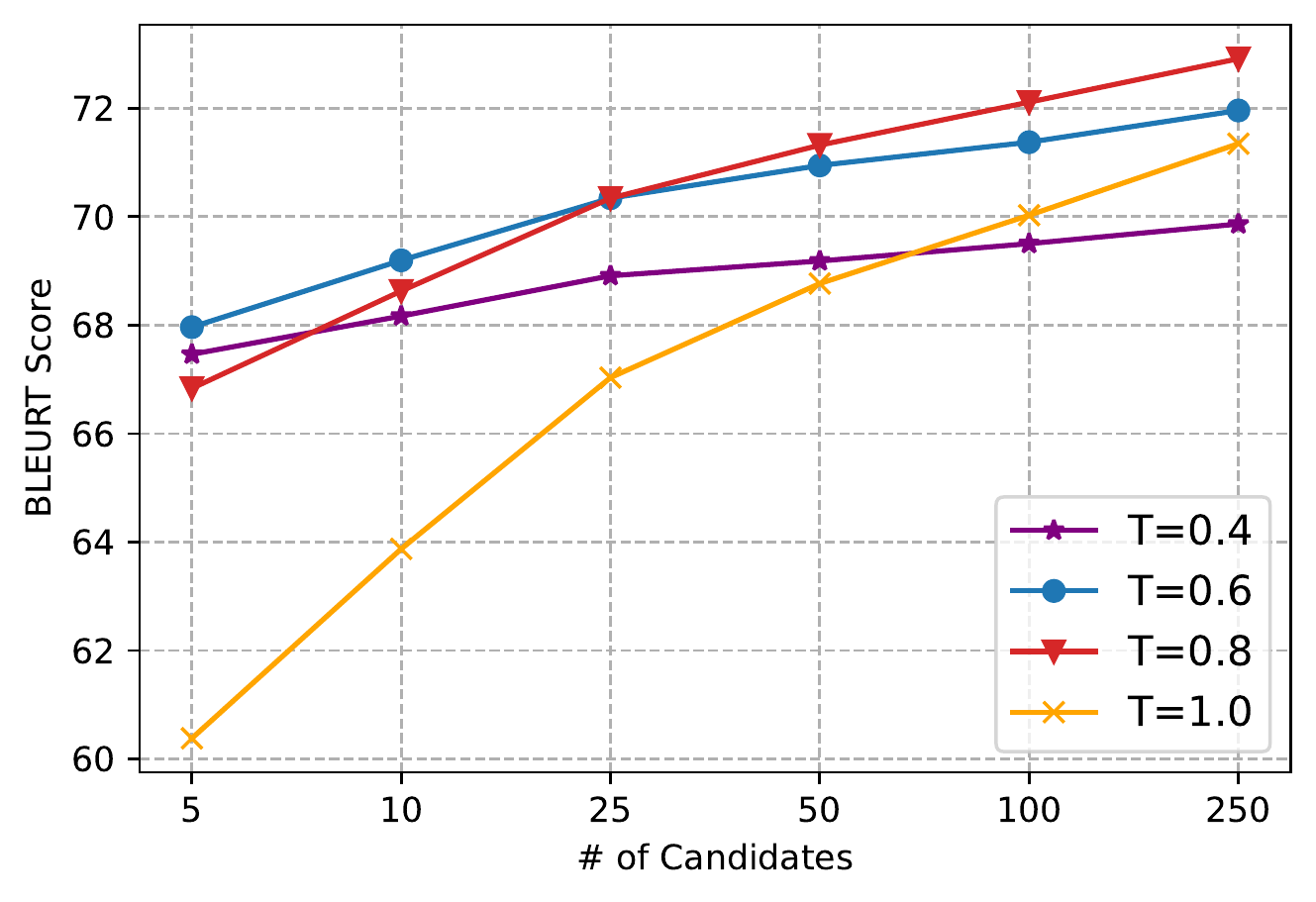}
  \label{fig:sub2}
\end{subfigure}
\vspace{-5mm}
\caption{BLEURT scores against the number of candidates used with different temperature values. All results are averaged over three random runs. \textbf{Left}: Transformer-base w/ LS=0.1; \textbf{Right}: Transformer-base w/o LS}
\label{fig:t_vs_k}
\vspace{-15pt}
\end{figure}

Figure \ref{fig:t_vs_k} shows the results. Given the model with label smoothing, distributional cooling is necessary. Even with 250 candidates sampled per sentence, our methods ($T<1.0$) strongly outperform the naive MBR ($T=1.0$) by a considerable margin. 
Given the model without label smoothing, our methods still significantly outperform naive MBR in most cases, except in the scenario of high costs (e.g., $N$=250), where a proper choice of temperature (e.g., $T$=0.6/0.8) is required. 
For both models, our approach helps MBR achieve strong performance at a low cost ($N$=5,10). 
{\it The above results resolve the concern that sharpening model distribution would limit the gains with a large number of candidates.} 

In addition, we find that the performance of our approach improves with an increasing number of candidates, indicating our approach retains the advantage of MBR of not suffering from the \emph{beam search curse} problem~\cite{koehn-knowles-2017-six,eikema2021sampling}. 


\subsection{Tempearture of DC-MBR}
\label{sec:exp_hyp_and_ref_space}
Temperature is another key factor for DC-MBR.

\begin{figure}[t]
\centering
  \includegraphics[width=0.8\linewidth]{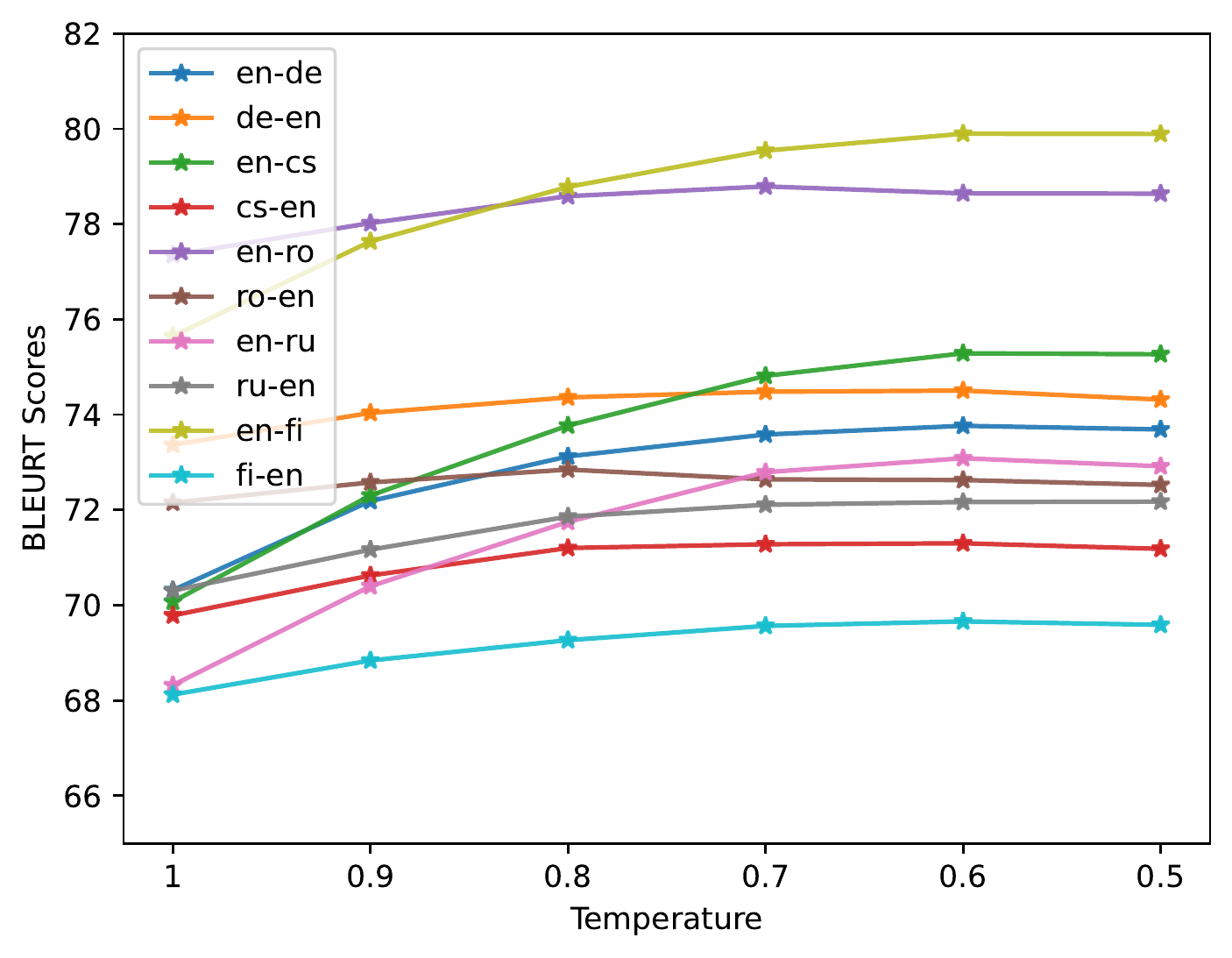}
\caption{BLEURT scores for varying DC-MBR's temperature values on multilingual tasks. We use $N=10$. Best viewed in color. }
\label{fig:multi_bleurt_vs_temp}
\vspace{-13pt}
\end{figure}

\noindent\textbf{\emph{Applicability of Distributional Cooling.}} Besides DC-MBR's effectiveness shown in previous experiments, we want to know whether DC-MBR is applicable to wider settings such as different translation directions. To this end, we plot the performance of each direction of our multilingual experiments in Figure \ref{fig:multi_bleurt_vs_temp}.
As shown, tuning down temperature almost monotonically improves translation performance in all directions, proving the general applicability of DC-MBR.

\noindent\textbf{\emph{Distinct Roles for $\mathcal{Y}_h$ and $\mathcal{Y}_r$.}}
In the above experiments, we use the same temperature to generate both the hypothesis and reference space. 
Since they have very different roles in MBR decoding, we study how the performances are affected by temperature.

\begin{figure}[t]
\centering
  \includegraphics[width=\linewidth]{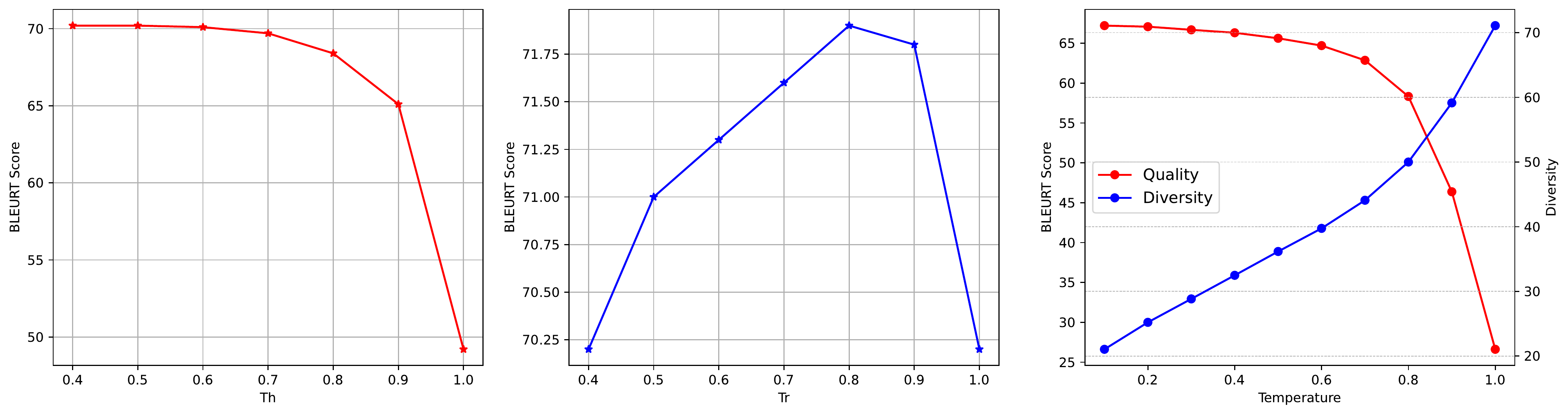}
\caption{Temperature study of DC-MBR. The number of candidates is 10. \textbf{Left}: Fix $T_r=1.0$ and tune $T_h$; \textbf{Middle}:  Fix $T_h=0.5$ and tune $T_r$. \textbf{Right}: Quality/Diversity scores of sampling candidates versus Temperature $T$. \it{Red}: Quality; \it{Blue}:  Diversity }
\label{fig:hyper_study}
\vspace{-13pt}
\end{figure}

Figure \ref{fig:hyper_study} shows the choices of $T_h$ and $T_r$ for the hypothesis and reference space, respectively. Experiments are conducted on the valid set of WMT20 En-De. The model we use is the Transformer-base trained with label smoothing 0.1.
As shown, $T_h$ has a significant effect on the performance of MBR. The BLEURT score gap between the best ($T_h=0.5$) and worst settings ($T_h=1.0$) is about 20 points. 
A sharp hypothesis space is more favorable than a smooth one. 
On the other hand, a good $T_r$ value also provides a considerable gain on the performance, about +1.5 BLEURT scores.
Different from $T_h$, a sharp reference space is not always the best choice. 
A $T_r$ value that is too high or too low can result in a drop in BLEURT.

In order to further reveal the characteristics of both spaces, the right plot in Figure \ref{fig:hyper_study} plots the quality and diversity for sampled candidates over different choices of temperature. We directly use the BLEURT score for quality, and the diversity score is defined as,
\begin{gather}
    \text{Div} = \frac{1}{|\mathcal{Y}|^2}\sum_{a\in \mathcal{Y}} \sum_{b\in \mathcal{Y}} \text{ChrF}(a, b),
\end{gather}
which is the average ChrF score~\cite{popovic2015chrf} of each candidate against the others. We do not use neural metrics such as BLEURT because they are trained to be robust against surface changes. 
In conclusion, the hypothesis space provides the possible candidates for translation, which should be of high quality and insensitive to diversity. In contrast, the reference space is responsible for the comprehensive evaluation of utilities, which should balance both quality and diversity.



\section{Conclusion}
We investigated the negative effect of label smoothing on MRB, finding that MBR's performance decreases monotonically with the increase of label smoothing value, and showing that the above phenomenon is due to the \emph{autoregressive over-smoothness} caused by the autoregressive factorization. 
We then presented a conceptually simple and theoretically well-motivated approach, DC-MBR, to address this issue.
Extensive experiments on NMT benchmarks revealed that the widely used \emph{un-biased} setting is sub-optimal, and demonstrated that our method significantly improves MBR.



\section{Limitations}

First, in this work, we focus on machine translation, as the MT community mainly witnesses MBR's reported success. Recently, researchers start to investigate minimum risk decoding with similar ideas with open-ended text generations \cite{suzgun2022follow}. In these scenarios, diversity is a crucial criterion, and sharpening the model's distribution may deteriorate the performance of these tasks.


Second, as the NLP community's growing attention on the large-scale foundation models~\cite{bommasani2021opportunities}, one limitation of this work is that it has not explored the variety of settings brought by these large models, with only considering mBART-50 for direct multilingual translation. 
For instance, recent models, like MASS~\cite{song2019mass}, GPT3~\cite{brown2020language}, and OPT~\cite{zhang2022opt}, have shown remarkable progress in multilingual few-shot translations or in-context learning for translations \cite{garcia2022using, agrawal2022context}. 
These possible scenarios of foundation models could fit well for DC-MBR and worth exploring. 



\bibliography{anthology,custom}
\bibliographystyle{acl_natbib}

\appendix

\include{appendix}

\end{document}

%% file: appendix.tex
\section{Derivation of Optimal Solutions}
Even though we solve the problem by gradient descent in practice, the label-smoothed cross-entropy objective has an optimal analytic solution.

Recall that the label smoothed cross-entropy loss is defined by,
\begin{gather}
    \mathbf{L}_{ls} = - \sum_{i} Q^i_{\lambda} \cdot \log {{P^i}}, \\
    Q^i_{\lambda} = \begin{cases}
      1-\lambda & \text{if $y^i$ is golden token}\\
      \frac{\lambda}{|V|-1} & \text{otherwise}
    \end{cases}.
\end{gather}
We short $P(y^i)$ as $P^i$ for simplicity.
Then, we can write the constrained optimization problem as,
\begin{align*}
\min_{P^i} \quad & - \sum_{i} Q^i_{\lambda} \cdot \log {{P^i}} \\
\textrm{s.t.} \quad & \sum_{i}P^i = 1
\end{align*}
Then, we can add a constant $\sum_i Q^i_{\lambda} \cdot \log Q^i_{\lambda}$ to the objective, which will not affect the choice of solution $\hat{P}$,
\begin{align*}
\min_{P^i} \quad & \sum_{i} Q^i_{\lambda} \cdot \frac{\log Q^i_{\lambda}}{\log {{P^i}}} = \textrm{KL}(Q_{\lambda}, P) \\
\textrm{s.t.} \quad & \sum_{i}P^i = 1
\end{align*}
Obviously, to achieve the minimum (i.e., 0) of KL divergence, two distributions $P$ and $Q$ must be the same. Hence, the $i$-th token's probability value of the optimal solution $\hat{P}$ equals to,
\begin{align}
    \hat{P}^i_{\lambda} = \begin{cases}
      1-{\lambda} & \text{$y^i$ is golden token} \\
      \frac{{\lambda}}{|V|-1} & \text{otherwise}
    \end{cases}.
\end{align}

\section{Derivation of Equivalence between Label Smoothing and Distributional Cooling}
Here, we provide detailed proof for the equivalence of post-tuning temperature and pre-tuning label smoothing factor.
Give two models trained with label smoothing factor $\lambda_1, \lambda_2 < 1$.
Based on Proposition \ref{prop:optimal_solution}, the optimal solutions are 
\begin{gather*}
    \hat{P}_{\lambda_1}=[\frac{{\lambda_1}}{|V|-1}, \cdots, 1-{\lambda_1}, \cdots,  \frac{{\lambda_1}}{|V|-1}], \\
    \hat{P}_{\lambda_2}=[\frac{{\lambda_2}}{|V|-1}, \cdots, 1-{\lambda_2}, \cdots,  \frac{{\lambda_2}}{|V|-1}],
\end{gather*}
We want to see whether there exists a temperature value $T$ which can transform $\hat{P}_{\lambda_1}$ to $\hat{P}_{\lambda_2}$.

First, with the Softmax formulation, we have,
\begin{gather}
    \frac{\exp{o_{gt}}}{\sum_j \exp{o_j}} = 1-\lambda_1,
\end{gather}
where $o_{gt}$ is the logit for the golden target token.
It is obvious that the non-target tokens share the same logit value, which we refer to as $o_{nt}$.
Then, we have
\begin{align*}
    \quad &\frac{\exp{o_{gt}}}{\sum_j \exp{o_j}} = 1-\lambda_1 \\
    \Longleftrightarrow \quad & \frac{\exp{o_{gt}}}{\exp{o_{gt} + (|V|-1) \cdot \exp{o_{nt}} }} = 1-\lambda_1 \\
    \Longleftrightarrow \quad & \exp{o_{gt}} = \frac{1-\lambda_1}{\lambda_1} \cdot (|V|-1) \cdot \exp{o_{nt}}
\end{align*}

Therefore, if we can find a solution $T$ that satisfies both,
\begin{gather}
    \label{eq:1}
    \exp{o_{gt}} = \frac{1-\lambda_1}{\lambda_1} \cdot (|V|-1) \cdot \exp{o_{nt}}\\
    \label{eq:2}
    \exp{\frac{o_{gt}}{T}} = \frac{1-\lambda_2}{\lambda_2} \cdot (|V|-1) \cdot \exp{\frac{o_{nt}}{T}},
\end{gather}
we obtain the optimal solution for label smoothing $\lambda_2$.
By taking the logarithm of both equation \ref{eq:1} and \ref{eq:2}, we obtain,
\begin{align*}
    &\begin{cases}
      o_{gt} &= \log \frac{(1-\lambda_1)(|V|-1)}{\lambda_1} + o_{nt} \\
      o_{gt} &= T \cdot \log \frac{(1-\lambda_2)(|V|-1)}{\lambda_2} + o_{nt} \\
    \end{cases}, \\
    \Rightarrow & \log \frac{(1-\lambda_1)}{\lambda_1} = T \cdot \log \frac{(1-\lambda_2)}{\lambda_2}, \\
    \Rightarrow & T = (\log \frac{1-\lambda_1}{\lambda_1})/(\log \frac{1-\lambda_2}{\lambda_2}).
\end{align*}
Thus, for any given $\lambda_1$ and $\lambda_2$, we can have a temperature $T$ that can transform the optimal solution for label smoothing $\lambda_1$ to $\lambda_2$.

\begin{table*}[]
\centering
\begin{tabular}{c|c|c|c|c|c}
\toprule
\textbf{Utility $\backslash$ Evaluation} & \textbf{BLEU} & \textbf{ChrF}	& \textbf{Meteor} & \textbf{BEER}	& \textbf{BLEURT} \\ \midrule
\textbf{BLEU} & \textbf{29.0 / 27.1} & 56.8 / 54.6 & 38.8 / 37.0 & 60.7 / 59.4  & 72.0 / 68.8 \\ \midrule
\textbf{ChrF} & 28.7 / 25.5 & \textbf{57.3 / 55.7} & 38.9 / 37.0 & 60.7 / 59.3  &72.2 / 69.4 \\ \midrule
\textbf{Meteor} & 28.6 / 25.4 & 56.9 / 54.8 & \textbf{39.0 / 37.6} & 60.6 / 59.1 & 72.0 / 69.1 \\ \midrule
\textbf{BEER} & 29.0 / 26.5 & 57.0 / 55.3 & 38.8 / 37.2 & \textbf{60.9 / 59.9} & 72.0 / 69.5\\ \midrule
\textbf{BLEURT} & 28.7 / 23.9 & 56.9 / 53.4 & 38.7 / 35.4 & 60.6 / 57.9 & \textbf{75.0 / 74.5}\\ 
\bottomrule
\end{tabular}
\caption{Results of different utility functions and evaluation metrics. We report the average score of scores in ten directions of WMT16. Each cell contains results of DC-MBR at the left and naive MBR at the right, split by a slash symbol. }
\label{tab:util_funcs}
\end{table*}

\begin{table}[th]
\centering
\resizebox{\columnwidth}{!}{%
\begin{tabular}{l|l|c|c|c}
\toprule
& Dataset     & Train & Valid & Test \\
\midrule
\multirow{3}{*}{Bilingual} & WMT20 En-De & 37M   & 1997  & 1418 \\ 
& WMT20 De-En & 37M   & 2000  & 785 \\
& WMT16 En-RO & 608K & 1999 & 1999 \\
\midrule
\multirow{10}{*}{Multilingual} & WMT16 En-De & -   & 2169 & 2999 \\ 
& WMT16 De-En & -   & 2169 & 2999 \\ 
& WMT16 En-Cs & -   & 2656 & 2999 \\ 
& WMT16 Cs-En & -   & 2656 & 2999 \\ 
& WMT16 En-Ro & -   & 1999 & 1999 \\ 
 & WMT16 Ro-En & -   & 1999 & 1999 \\ 
 & WMT16 En-Ru & -   & 2818 & 2998 \\ 
 & WMT16 Ru-En & -   & 3003 & 2998 \\ 
 & WMT16 En-Fi & -   & 1500 & 3000 \\ 
 & WMT16 Fi-En & -   & 1500 & 3000 \\ 
\bottomrule
\end{tabular}}
\caption{Dataset statistics.}
\label{tab:data_stat}
\vspace{-10pt}
\end{table}

\section{Detailed Experimental Setup}
\label{detail_exp_setup}
In this section, we provide detailed experimental settings that help the readers to reproduce our results. 

For the bilingual NMT setting, we train models from scratch and preprocess datasets following previous work. 
For En-De and De-En, We apply the same filtering process described in~\citet{zeng2021wechat} and get about 37M parallel sentence pairs, which are tokenized with Moses \footnote{http://www.statmt.org/moses/} and segmented by byte pair encoding BPE~\cite{sennrich-etal-2016-neural} with 32000 merge operations. 
For En-Ro, we have 608k parallel sentences tokenized and segmented using the same tool as En-De and De-En. 

For the multilingual NMT setting, we use the large version of the released mBART-50\footnote{\url{https://huggingface.co/facebook/mbart-large-50}}. The benchmarks we used are the ten tasks of WMT16 supported by mBART-50. 
The dataset statistics of both bilingual and multilingual settings can be found in Table \ref{tab:data_stat}.

For the significance test, we first tokenize the generated sequence and reference with the tokenizer\footnote{\url{https://github.com/moses-smt/mosesdecoder/blob/master/scripts/tokenizer/tokenizer.perl}}, and then use the script provided by the Moses toolkit\footnote{\url{https://github.com/moses-smt/mosesdecoder/blob/master/scripts/analysis/bootstrap-hypothesis-difference-significance.pl}}. It is worth noting that the significance test is conducted with NIST \cite{nist} and BLEU \cite{papineni2002bleu} scores.

\section{More Bilingual NMT Results}
\label{sec:more_bi_exps}
This section provides experiments on WMT16 En-Ro and WMT20 De-En under the bilingual NMT setting. 
The results are shown in Table \ref{tab:en-ro} and Table \ref{tab:de-en}. We can see that the results are consistent with the conclusions in the main text, justifying the generalization of our method.

\begin{table}[t]
\centering
\begin{subtable}{.5\textwidth}
\small
\centering
\begin{tabular}{c|l|c|c|c}
\toprule
\textbf{Models} & \multicolumn{1}{c|}{\textbf{Models}} & \textbf{BS} & \textbf{MBR} & \textbf{Ours} \\ \midrule
\multirow{4}{*}{\textbf{$N=10$}} & Transformer & 29.1 &28.9	&28.9 \\ \cline{2-5} 
                               & \cellcolor{gray!20}~~ + LS 0.1    &  31.1 & 25.1	& 30.9 \\ \cline{2-5} 
                               & \cellcolor{gray!20}~~ + LS 0.2    &  31.4 & 19.7	& 31.2 \\ \cline{2-5} 
                               & \cellcolor{gray!20}~~ + LS 0.3    &  31.5 & 12.5	& 30.9 \\ \midrule
\multirow{4}{*}{\textbf{$N=50$}} & Transformer & 29.1 &  29.4	& 29.1  \\ \cline{2-5} 
                               & \cellcolor{gray!20}~~ + LS 0.1    &  31.1 & 27.7	&31.2 \\ \cline{2-5} 
                               & \cellcolor{gray!20}~~ + LS 0.2    & 31.4 & 22.5	& 31.5 \\ \cline{2-5} 
                               & \cellcolor{gray!20}~~ + LS 0.3    & 31.5 & 15.8	& 31.3 \\ \bottomrule
\end{tabular}
\caption{SacreBLEU scores on WMT16 En-Ro task.}
\label{tab:en-ro}
\end{subtable}
\begin{subtable}{.5\textwidth}
\centering
\small
\begin{tabular}{c|l|c|c|c}
\toprule
\textbf{Models} & \multicolumn{1}{c|}{\textbf{Models}} & \textbf{BS} & \textbf{MBR} & \textbf{Ours} \\ \midrule
\multirow{4}{*}{\textbf{$N=10$}} & Transformer & 70.5 & 66.6 & 71.5 \\ \cline{2-5} 
                               & \cellcolor{gray!20}~~ + LS 0.1    & 70.5 & 49.4 & 71.2 \\ \cline{2-5} 
                               & \cellcolor{gray!20}~~ + LS 0.2    & 70.2 & 39.2 & 71.3 \\ \cline{2-5} 
                               & \cellcolor{gray!20}~~ + LS 0.3    & 70.6 & 33.2 & 71.2 \\ \midrule
\multirow{4}{*}{\textbf{$N=50$}} & Transformer & 70.5 & 69.9 & 72.0 \\ \cline{2-5} 
                               & \cellcolor{gray!20}~~ + LS 0.1    & 70.5 & 55.4 & 71.9 \\ \cline{2-5} 
                               & \cellcolor{gray!20}~~ + LS 0.2    & 70.2 & 43.7 & 71.9 \\ \cline{2-5} 
                               & \cellcolor{gray!20}~~ + LS 0.3    & 70.6 & 37.6 & 71.9 \\ \bottomrule
\end{tabular}
\caption{BLEURT scores on WMT20 German-English Task. }
\label{tab:de-en}
\end{subtable}
\end{table}

\section{Study of Utility Functions}
\label{sec:util_and_eval}
In the main text, we report results with BLEURT as our utility function following \citet{freitag2021minimum}. 
In this section, we provide results using two more metrics, i.e., sacreBLEU and ChrF, as our utility function. 
The experiments are conducted in the multilingual NMT setting.
We also include BLEURT results for comparison. 

Table \ref{tab:util_funcs} shows the results. We mark the best-performed result of each evaluation metric as bold. As shown, the same utility function yields the best performance with the corresponding evaluation metric, which is consistent with findings of \citet{freitag2021minimum}. Then, the performance of DC-MBR exceeds that of naive MBR over all utility functions and all evaluation metrics, with considerable margins. The above results prove the compatibility of DC-MBR.